# Ontology-Based Post-Coordinated Expression Assembly from Clinical Narrative into Queryable Knowledge Graphs

## Authors and affiliations

Johann PIGNAT*[1,2], Christophe GAUDET-BLAVIGNAC[1], Julien EHRSAM[1], Jean-Philippe GOLDMAN[1], Cyrille DURET[1], Phil CHENG[2], Mina BJELOGRLIC[1], Michel CUENDET[2], Christian LOVIS[1], Olivier MICHIELIN[2]

*Corresponding author (johann.pignat@hug.ch)

Affiliations:

1) Service des sciences de l'information médicale, Hôpitaux Universitaires de Genève, Geneva, Switzerland
2) Service d'oncologie de précision, Hôpitaux Universitaires de Genève, Geneva, Switzerland

## Abstract

Clinical oncology documentation records decisive information in free text, where it remains readable but not retrievable at scale, leaving cohort identification dependent on incomplete diagnostic codes or manual review. The current standard, coding individual terms, captures what a document mentions, not what it asserts: meaning lies in how those terms relate to one another. We present a method for deriving queryable knowledge graphs from clinical narratives, conforming to the Extended Grammar of SNOMED CT across multiple ontologies. The method specifies how entities should be annotated and related. It defines five relation classes, each bound to a construction rule that assembles the entities and relations into post-coordinated expressions. A deterministic implementation, UNIGRAPHE, applies those rules to construct knowledge graphs from annotated text. Applied to expert-annotated real-world oncology documentation, UNIGRAPHE reconstructed 5,299 annotations into 2,196 post-coordinated expressions and shows that relation structure is the missing piece to make clinical narrative formally queryable.

# Introduction

Clinical oncology documentation captures diagnostically and prognostically critical information (including metastasis topology, treatment response, and progression status) predominantly in free-text narrative. These documents are generated routinely and at scale yet remain computationally inaccessible for most research purposes: readable but not queryable by the formal mechanisms applied to structured data[1]. Transforming the narrative record into machine-queryable knowledge representations requires more than accurate entity extraction; it requires a representation architecture that clinical information systems can interrogate directly.

Clinical natural language processing (NLP) applications to oncology and general clinical text have grown substantially in recent years[2,3]. A substantial share of this work focuses on extraction and classification of discrete clinical items directly from narrative text: recent scoping reviews document extensive use of large language models for extracting patient and tumour characteristics from oncology records[4]. Dedicated pipelines recognise and classify clinical entities such as symptoms directly from unstructured notes[5]. A parallel strand of work instead uses free text to train models on the same narrative content to predict future clinical outcomes: progression-free survival[6], overall survival in metastatic disease[7], and, in critically ill populations more broadly, short- and long-term mortality[8]. Together, these results show that automated extraction from clinical narrative is increasingly mature; what its output still lacks is a semantic structure that can be formally queried to enable the narrative record to be reused, not one prediction at a time, but as a general resource for cohort discovery, computable phenotyping, and retrospective research at scale. Unlike single-endpoint, black-box predictive models, a compositional graph representation exposes each inference step and supports arbitrary downstream queries beyond the task it was originally designed for.

Gaudet-Blavignac et al.[9] formalised the grammar such a representation must conform to: a domain-general, ontology-grounded compositional grammar for SNOMED CT, extending its post-coordination model to integrate external ontologies and a relation vocabulary beyond its standard attribute set. We refer to this formalism as the Extended Grammar throughout this paper. What has been missing is a procedure for deriving post-coordinated expressions from a clinical narrative. The Extended Grammar defines the target, the method presented here defines how it is reached.

Other work has pursued graph representations of clinical data directly, organising extracted clinical text into relational or graph form rather than a flat concept list. Brens et al.[10] combine named entity recognition, terminology normalisation, and relation extraction to assemble FHIR-compliant patient records from free-text electronic health record (EHR) notes, explicitly capturing structured associations between conditions, medications, and observations. Arsenyan et al.[11] use large language models to construct a biomedical knowledge graph directly from EHR notes, extracting clinical entities and their interactions within a defined disease domain. Both demonstrate that clinical narrative can be organised beyond a flat list of normalised concepts. Neither, however, targets a domain-general, ontology-grounded compositional grammar of the kind that has been formalised by the Extended Grammar: FHIR resource linkage does not capture the attribute-value refinement or situational context such a grammar requires, and knowledge graph construction of this kind is typically scoped to a fixed relation vocabulary within a single clinical domain, evaluated by extraction accuracy rather than validated for formal queryability.

In practice, the dominant paradigm applied to terminology coding remains named entity recognition followed by normalisation to standard terminology codes[5,12,13]. This produces flat concept identifier lists with no compositional structure. This is architecturally insufficient for SNOMED CT, the terminology underpinning most clinical knowledge infrastructure[14], whose expressive power rests on *post-coordination* due to: the assembly of a focus concept with explicit attribute-value refinements into a formally queryable expression. SNOMED CT concepts are either primitive (insufficiently defined to be distinguished from other concepts by their modelling relationships alone) or fully defined (their defining relationships are both necessary and sufficient to distinguish them from other concepts); post-coordinated expressions are fully defined by construction, since every attribute-value relationship is made explicit. Flat entity lists cannot be assembled into such a structure at all. Three compounding failure modes result from using pre-coordinated concepts only, as they would appear in a flat entity list (Figure 1): first, many clinically critical combinations lack a pre-coordinated SNOMED equivalent entirely[15–18]. Secondly, a substantial proportion of those that do exist are *primitive*: they lack formal attribute definitions and are therefore unretrievable by property queries[19]. Thirdly, even fully defined concepts may fail expected queries when their curated attribute values do not align with the hierarchical paths used in clinical search. Post-coordination, built from atomic components via explicit relations, addresses all three limitations.

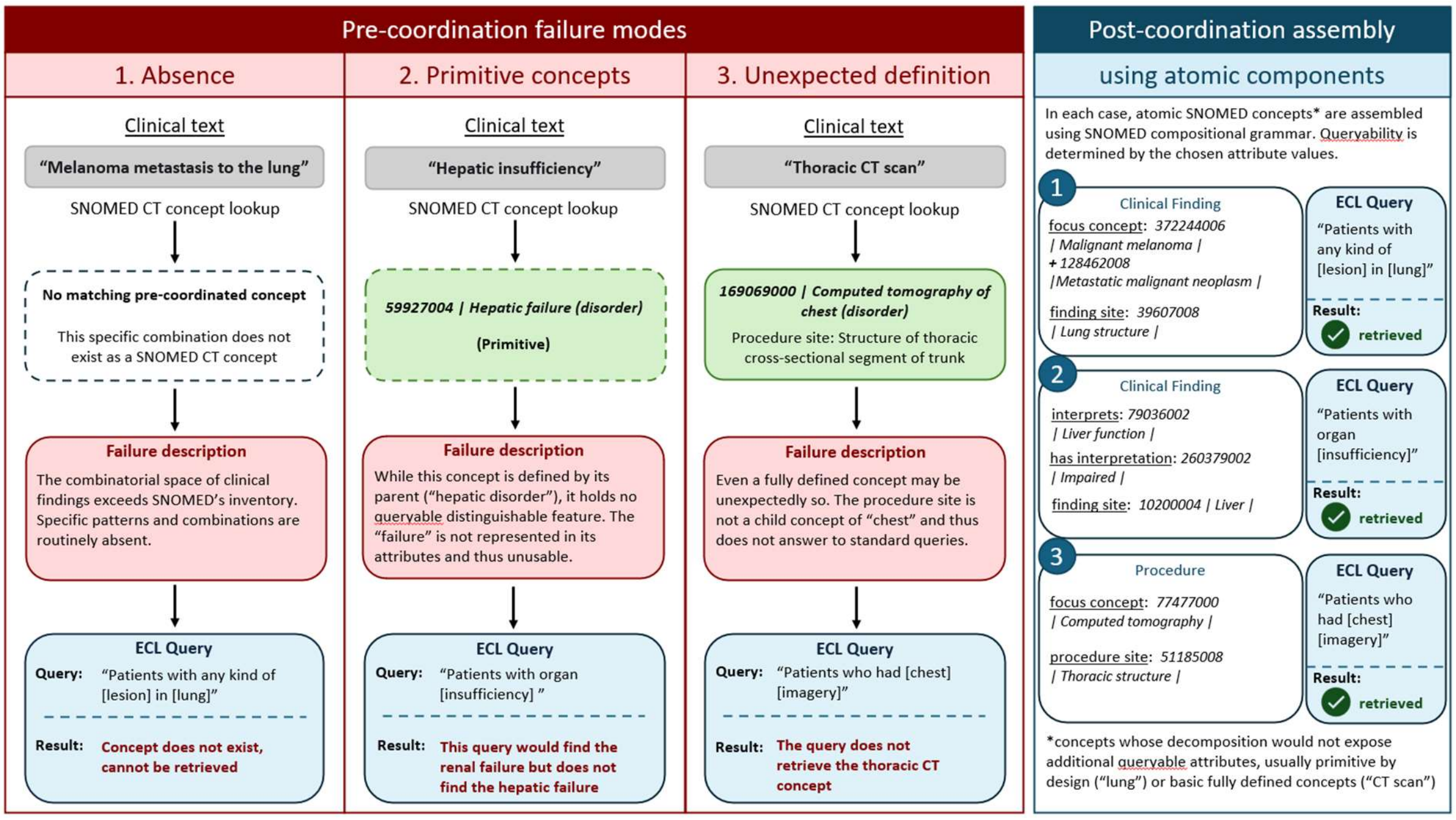

*Figure 1: Three independent failure modes of pre-coordinated SNOMED CT concept mapping, and their resolution by post-coordinated assembly*

Prior work toward automated post-coordination remains narrow in scope, and existing tooling largely assumes the expression already exists, a pattern confirmed by recent evaluations of automated SNOMED CT coding, which find that most systems normalise free text only to atomic, pre-coordinated concepts and do not construct expressions at all[20]. Kate[21] showed that individual clinical terms, including previously unlisted post-coordinated forms, could be converted into SNOMED CT concepts using a relation-identification model trained on existing terminology content, but the method operates on isolated terms rather than continuous narrative and targets a closed relation set rather than an extensible construction grammar. Peterson and Liu[22] extended automated relation identification to short free-text problem-list phrases using deep learning, and reported that relation identification between phrase and concept, rather than entity recognition, was the dominant remaining source of error, a finding consistent with the framing adopted here. Downstream of assembly, methods such as WASP[23], PCEtoFHIR[24], and ECLed[25] support the authoring, storage, and querying of post-coordinated expressions once they have been constructed, while Castell-Díaz et al.[26] use knowledge graph embeddings to predict a term's likely post-coordinated composition; none of these methods derives a complete expression, spanning situational context and inter-event structure, from an unconstrained clinical document. Existing clinical relation-extraction resources, including CACER for oncology events and relations[27], annotate relations between clinical entities without assigning them a construction role in a compositional target structure, a gap noted more broadly in recent reviews of clinical relation extraction[28]. The relation taxonomy presented here is intended to close this specific gap: it is not another method for recognising or normalising entities, but the missing specification for assembling them.

Realising post-coordination from free text requires a set of relations, classified according to a system that binds them each to one of a fixed set of construction rules. Each rule plays a distinct, non-substitutable role in the hierarchical assembly of valid post-coordinated expressions. Neither current clinical NLP systems nor existing annotation frameworks provide this specification. We present a reproducible method built on such a classification, comprising five classes empirically found sufficient to reconstruct any post-coordinated expression (detailed in Results). The application of the proposed method is demonstrated on a real-world oncology narrative corpus grounded in SNOMED CT and NCIt for oncology-specific entities, validated by expert clinical curators. We establish that the path from clinical narrative to formally queryable knowledge is, above all, a problem of relation structure.

The clinical stakes of the relation construction gap described above are concrete. Cohort identification for retrospective oncology research currently depends either on structured diagnostic codes, which capture only a fraction of the clinical reality documented in free text[29], or on manual record review, which does not scale[6]. A system capable of transforming free-text oncology documents into ECL-queryable knowledge graphs would make computable, for the first time, queries that currently require exhaustive manual review. Examples of such queries include: which patients had metastatic disease explicitly ruled out by a specific imaging modality, which patients showed disease progression on a particular treatment line, which patients were discussed at tumour board with a specific combination of findings. These are not edge cases, they are the routine building blocks of retrospective oncology research. This work extends our institution's existing knowledge graph in which structured EHR fields are already manually normalised using the Extended Grammar and queried using ECL; the aim is a parallel knowledge graph constructed from free-text clinical documents, grounded in the same ontological framework, enabling unified ECL-based cohort discovery across both structured and unstructured sources. We demonstrate this capability in Results through representative ECL queries against the reconstructed knowledge graph.

# Results

The primary outcome of this work is a reproducible method specifying how free text is annotated, as entities defined by an ontology code and connected by relations, and how those relations are then classified into one of five classes, each bound to a construction rule; together, the five rules allow reconstruction from single entities up to a complete post-coordinated expression. We first present the method, then its application to a real-world clinical corpus, and its qualitative validation.

## The Post-Coordinated Relation Taxonomy

We distinguish several related but separate terms used throughout this section:

- Annotated entity: a text span assigned an ontology normalisation code during the text annotation process; imported unchanged into the graph as a node.
- Annotated relation: a typed link drawn during annotation directly between two annotated entities; imported unchanged into the graph as an edge.
- Relation class: one of the five categories a relation must be assigned to.
- Construction rule: the operation bound to a relation class. Each rule takes nodes of specific types as input and produces new nodes of a specified type as output.
- Reconstructed node: the node a construction rule produces (a Qualified Entity, Composite Entity, Governing Concept, Situation, or Story).
- Reconstructed relation: a typed link a construction rule creates between nodes in the graph, typically inheriting the label of the annotated relation that triggered it, but connecting whatever nodes the rules have built by that point, not necessarily the original entities.
- Role (Component or Assertion): whether a node currently stands as a building block inside another node's construction (Component), or as the terminal representation of a complete clinical assertion (Assertion). A node's Assertion role can transfer to a new node built from it, for instance from a Governing Concept to the Situation that wraps it, at which point the original becomes a Component
- UNIGRAPHE: the deterministic pipeline that applies the construction rules to the imported graph to produce the reconstructed expression

Annotated relations are drawn directly between annotated entities  in BRAT and form a flat, unordered list of pairwise named links with no inherent hierarchy. UNIGRAPHE first imports the annotated entities and relations, exactly as recorded during annotation, into the graph. It then applies the five construction rules, one relation class at a time, to build the post-coordinated expression. Each relation Class is bound to exactly one Rule. The five Rules are always applied in a fixed order: *Qualification, Composition, Instantiation, Contextualisation, Connection*. Each Rule has defined input node types and outputs a specific node type (Table 1). A Rule's input is not always the original Annotated Entity: because earlier Rules may already have refined or combined entities, each reconstructed relation's endpoints are resolved against the current graph state before a Rule acts on them (see Methods for the resolution mechanism). At each step of the UNIGRAPHE pipeline, the Rules are applied deterministically based on annotated relation label and node type: given a fixed set of annotated entities and relations, the same input always produces the same output graph.

| **Node type** | **Role** | **Produced by** | **Input** | **Trigger Annotated Relation (examples)** |
|---|---|---|---|---|
| Annotated Entity | Component | Annotator | BRAT span + normalisation code | - |
| Qualified Entity | Component | Rule I – Qualification | ≥1 Annotated or Qualified Entity | Class I (laterality, direct_qualifier) |
| Composite Entity | Component | Rule II – Composition | Two same-level Entities | Class II (complement) |
| Governing Concept | Assertion, unless wrapped in a Situation (Component) | Rule III – Instantiation | ≥1 Entity (of any type) | Class III (finding_site, procedure_site, clinical_course) |
| Situation | Assertion | Rule IV – Contextualisation | One Governing Concept + explicit-context descriptor in the text | Class IV (finding_context, temporal_context) |
| Story | - | Rule V – Connection | ≥2 nodes holding the Assertion role | Class V (has_focus, due_to, finding_method) |

Table 1: Node types and the five Rules producing them from the five Classes of annotated relations.

In the following, we describe each relation Class and the associated Rule in the UNIGRAPHE pipeline, using the sentence “Biopsy for suspected severe malignant melanoma of the skin of the left leg" to illustrate the procedure. Figure 2 shows the sentence with its original Annotated Entities (in BRAT) and provides examples of each relation Class taken from this sentence. Figure 3 displays the final hierarchical knowledge graph produced and the links to the underlying Annotated Entities.

**Class I → Rule I: Qualification.** Rule I absorbs one Entity into another: the absorbed Entity's meaning folds into the entity it qualifies, producing a single Qualified Entity. The input can be either Annotated Entities or previously Qualified Entities. Rule I is applied to all Class I relations, before rule II begins. A qualifier ("*direct_qualifier*") relation narrows the qualified concept to a more specific instance of the base concept (e.g. "left" + "leg" → "left leg"). A *modifier* relation unlike a qualifier, does not narrow the entity into a more specific instance of itself: the resulting Qualified Entity is a distinct, related Entity, not a subtype of the one it was built from (e.g. "peri-" + "hepatic" → perihepatic, a location beside the liver, not part of it). The same pattern can also produce a new Qualified Entity under a different node label, connected back to the original entity by a reconstructed relation: a Substance absorbed into a Medicinal Product, or a Value absorbed into a Quantity).

**Class II → Rule II: Composition.** A Class II Relation (a "*complement*" relation) connects two Entities of equal conceptual standing. Rule II combines them into a single Composite Entity via an intersection (AND) operation, so neither is subordinate to the other. Two input nodes need not necessarily be at the same construction stage. In the running example, "skin" (a bare AnnotatedEntity) intersects with "left leg" (already refined into a Qualified Entity by Class I) into "skin of left leg," producing the composite anatomical argument required for Class III assembly.

**Class III → Rule III: Instantiation.** Rule III produces a new Governing Concept, the first node to hold the Assertion role, standing as a complete clinical statement rather than a building block for further assembly. Rule III is usually triggered by an Annotated Relation whose endpoints have already been resolved by Class I or II Rules. In some cases a single entity's own type can be enough on its own ("*haemorrhage*" as a Morphological Abnormality Entity, can be reconstructed as the corresponding Finding Governing Concept). Exactly one Governing Concept is produced by firing (Table 1); it holds the Assertion role by default, which may transfer to a Situation later via Rule IV. Relation labels are drawn from the SNOMED CT attribute set (e.g. finding site, associated morphology, procedure site), with a small number of custom additions covering natural-language patterns the standard attributes don't address (listed in Supplementary Table 2).

**Class IV→ Rule IV: Contextualisation.** Where the source text carries an explicit contextual descriptor overriding SNOMED CT's soft default context (i.e. the finding or procedure is not straightforwardly present, current, and about the patient, for example "suspected", "denies", "family history of"), a Class IV AnnotatedRelation connects that descriptor to the expression's Governing Concept resolved by Rule III; Rule IV wraps the Governing Concept in a Situation node, following the standard SNOMED CT Situation with Explicit Context pattern (e.g. "suspected" + malignant melanoma [Governing Concept] → Situation: "suspected malignant melanoma"). The Situation embeds the Governing Concept as the value of an *associated_finding* or *associated_procedure* attribute, together with the relevant reconstructed context attributes (certainty, temporality, or subject). The Assertion role, held by the Governing Concept since Rule III, transfers to the new Situation node; the Governing Concept itself remains in the graph, still addressable through the Situation's embedding attribute, but no longer represents the full expression on its own. Where no such descriptor is present, no Class IV Annotated Relation exists and the Governing Concept keeps the Assertion role, standing as the expression's terminal node in its own right.

**Class V → Rule V: Connection.** Rule V is resolved last, once both entities connected by a Class V Annotated Relation have been independently assembled into their own expressions, either a Governing Concept or a Situation, holding the Assertion role. Rule V connects these two Assertion-role nodes into a Story: a document-scoped subgraph representing the narrative connections between clinical assertions found explicitly in the source text (e.g. a biopsy Situation connected to the melanoma Situation it investigates). Stories are limited to the document in which they are constructed. By operating at this inter-Assertion level rather than embedding associative links inside individual expressions, Rule V preserves graph navigability and ECL-queryability while retaining the full narrative structure of the source document.

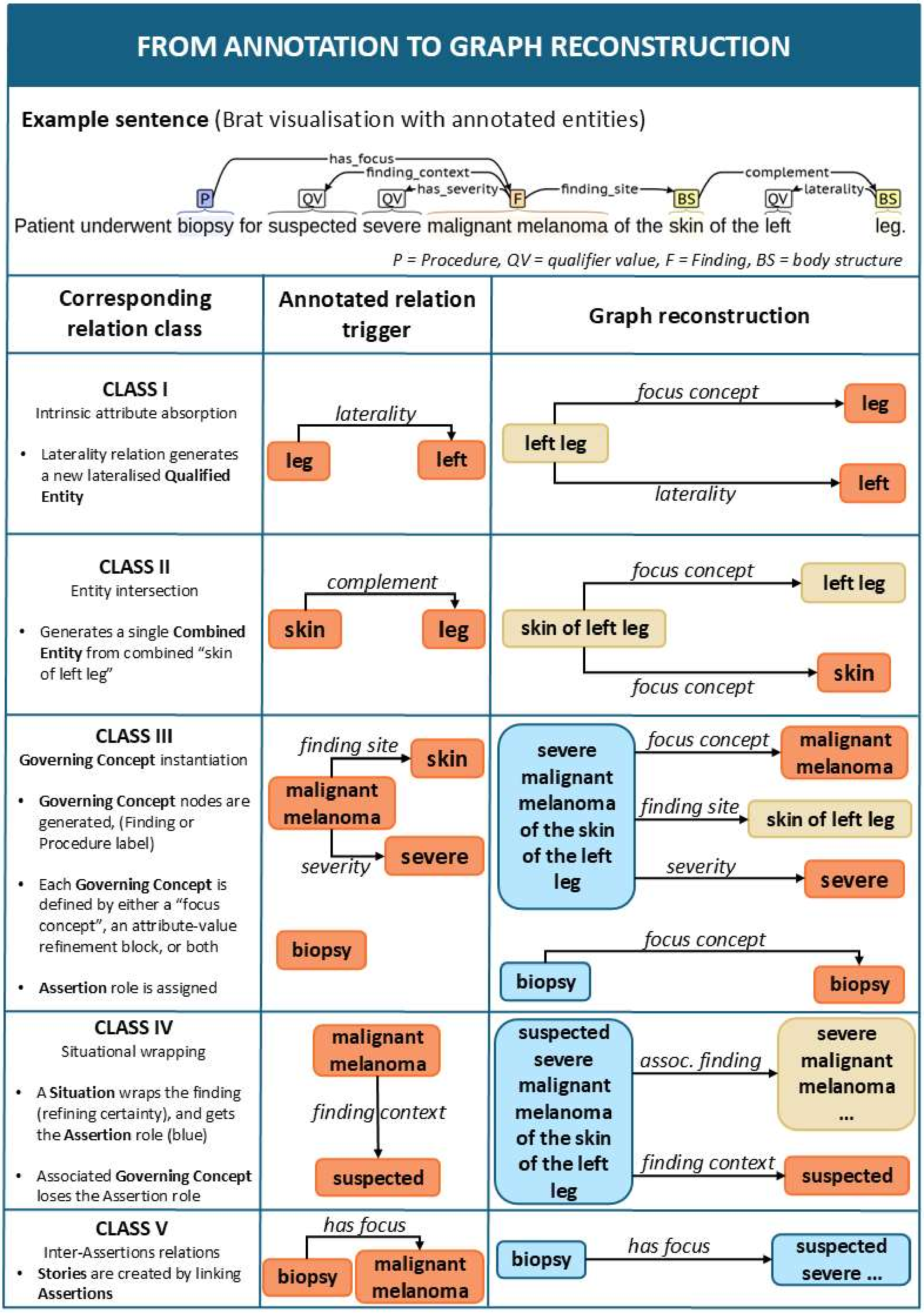

*Figure 2:* ***Relation class examples.*** *The five relation Classes and corresponding rules applied sequentially by the UNIGRAPHE construction pipeline, demonstrated on the running example sentence. Node colours: orange = Annotated Entity; beige = Qualified Entity / Composite Entity; blue = Assertion. Colours match Figure 4, which additionally shows provenance/metadata nodes.*

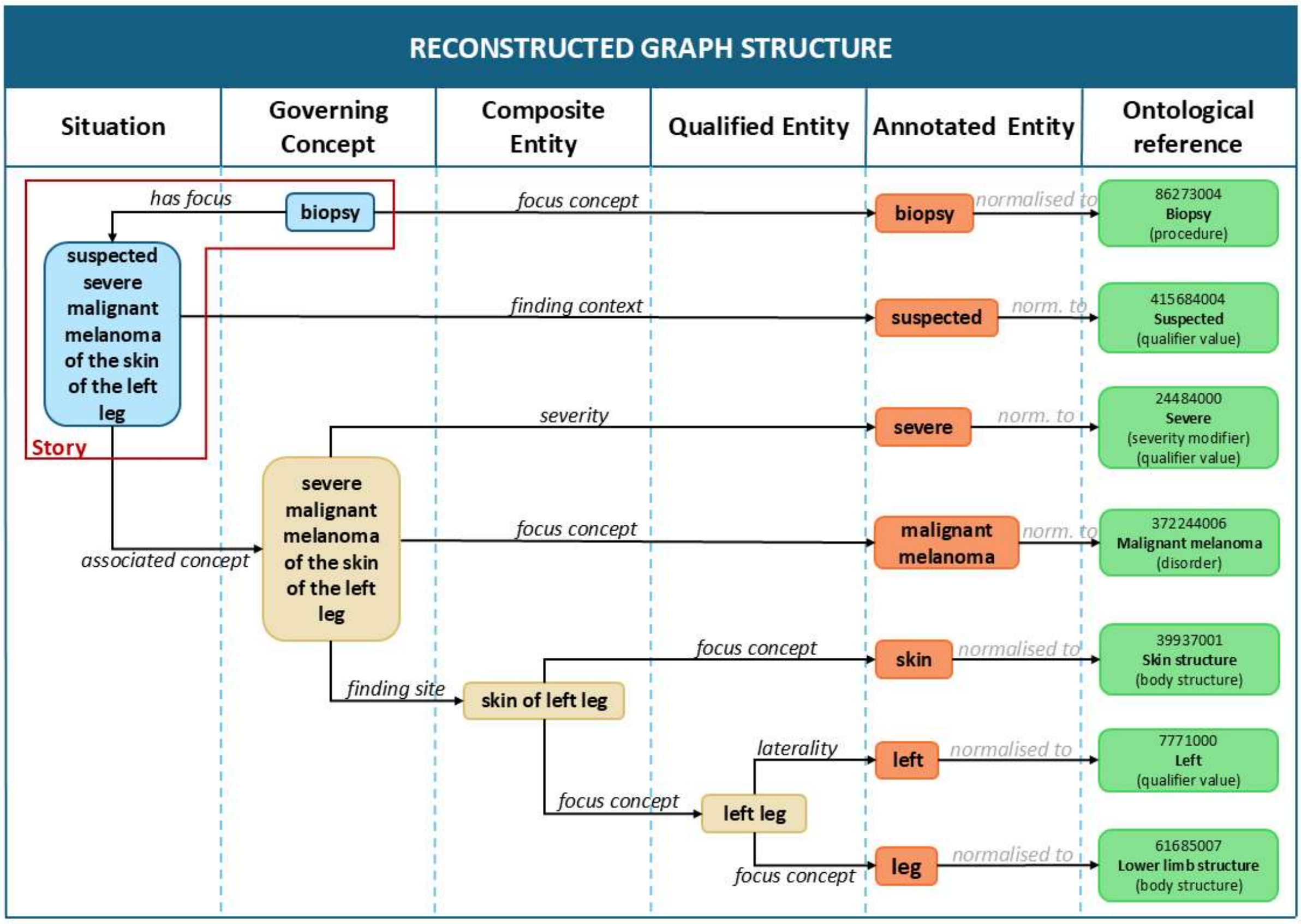

*Figure 3: Complete reconstructed knowledge graph assembled from the sentence "Biopsy for suspected severe malignant melanoma of the skin of the left leg" following execution of all five relation classes.*

## Knowledge graph architecture

The UNIGRAPHE reconstruction pipeline processes Annotated Entities and Relations through successive application of the five Rules to build the increasingly elaborate expressions as post-coordinated concepts, as described in Table 1. Every expression thus forms a node in a well-defined knowledge graph. Each such node retains an explicit traceability link to its originating text span, the original document it stems from, and the patient it relates to. This layered structure of clinical entities linked to their ontological references, both anchored to document and patient provenance, is what makes the reconstructed graph auditable end to end, a requirement for clinical and regulatory use (Figure 4).

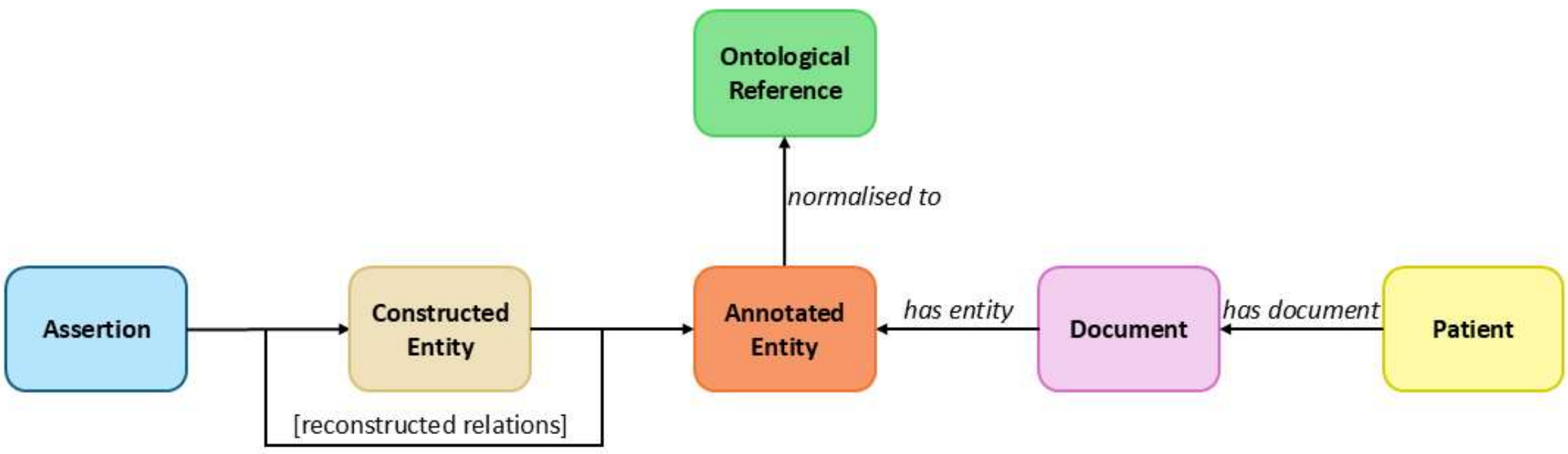

*Figure 4: Knowledge graph organisation, with text representation (Assertion, Entities), ontological reference and metadata (document/patient provenance).*

The node hierarchy explicitly mirrors the Extended Grammar: where present, a *focus concept* edge points to the node acting as "parent concept" of a Governing Concept or an Entity, while attribute edges (e.g. *finding site, associated morphology*) add the specific details that distinguish it from that parent. This structural alignment enables unified querying: any ECL query structured around attribute constraints operates identically across pre-coordinated structured data and free-text-derived post-coordinated expressions without modification to query syntax (Figure 5).

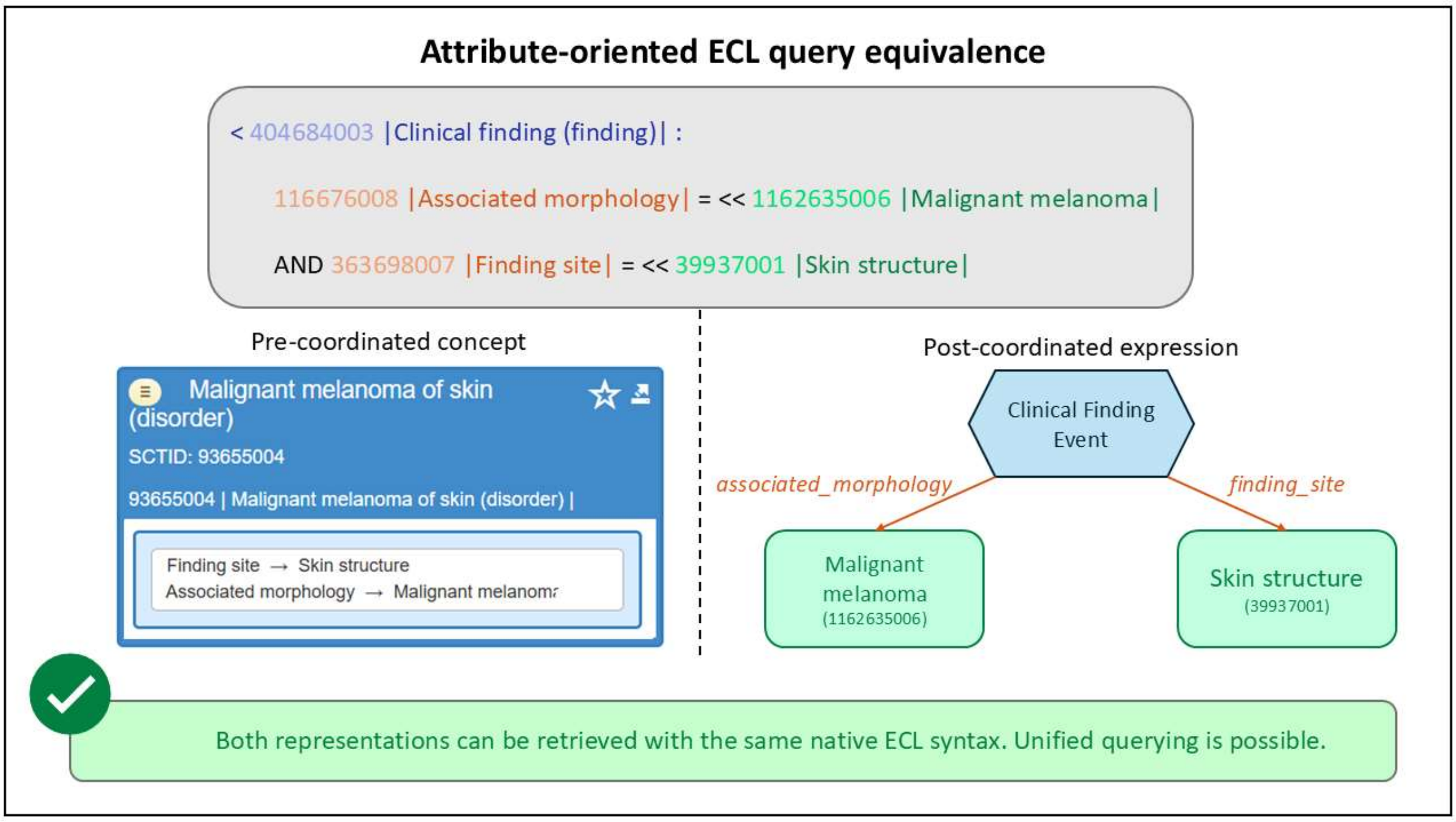

*Figure 5: Compatibility between the attribute-centred reconstruction approach and pre-coordinated normalisation for ECL queries. Left shows the pre-coordinated concept as presented in the SNOMED CT browser, which shows the same refinement block as the post-coordinated expression on the right.*

When source text provides no explicit focus concept for an assembled expression (a common occurrence when a clinical examination is described only by its interpretation, e.g. "normal exam") the pipeline assigns the appropriate high-level SNOMED root (Clinical finding or Procedure) as a default. Attribute-value refinements are fully preserved and queryable; only the hierarchical placement of the expression within SNOMED is underspecified. For attribute-based cohort discovery, the dominant clinical research use case[30,31], this approach is preferable to assigning a primitive pre-coordinated concept whose label is more specific but whose attributes are unexposed. Full details of the graph reconstruction algorithm, including Cypher query templates and JSON configuration files, are provided on the publication repository (see Code availability).

## Multiplicity handling

When a single textual entity serves as the anchor for multiple distinct governing concepts (e.g. "metastases of the lung and liver"), a Split Calculus mechanism duplicates the Governing Concept node, producing structurally independent but semantically parallel expressions. When an annotator intends to express a compound event as a single assertion, a group relation directs the algorithm to cluster targets into a unified Composite Entity before Class III assembly, bypassing the split logic. This logic is presented in Figure 6.

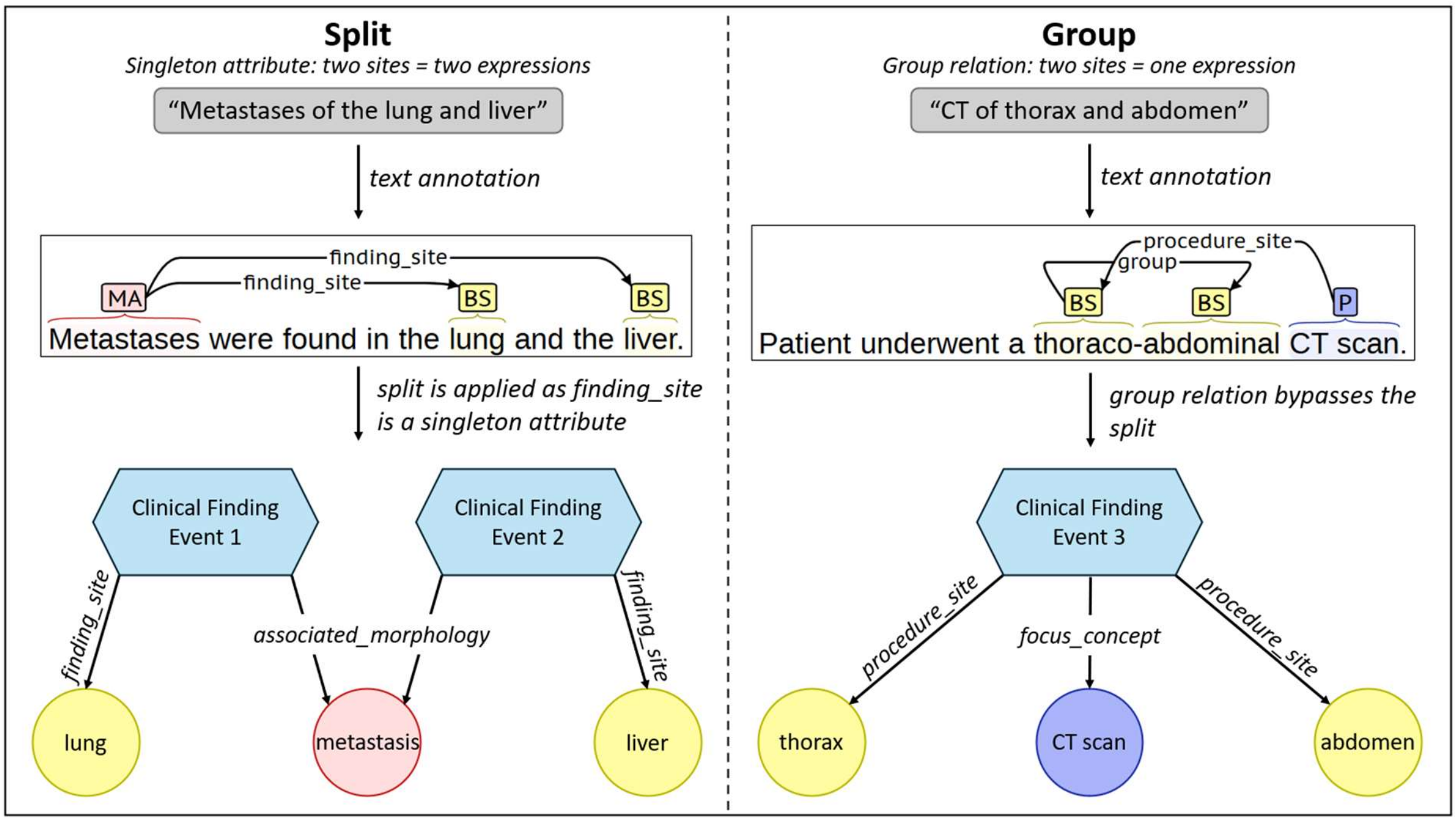

*Figure 6: The split mechanism governing multiplicity of Governing Concepts (blue). P = Procedure (purple); BS = Body Structure (yellow); MA = Morphologic Abnormality (red)*

## Corpus application

The corpus comprises 20 manually annotated documents from five purposefully selected patients (anonymised here as P1 through P5) across four document types, containing a total of 17,809 words. The pipeline deterministically reconstructed these annotations into 2,196 post-coordinated expressions from 5,299 manual annotations, orchestrated with a total of 4,273 manually annotated relations, 3,855 of which belong to one of the five relation classes (Figure 7). The full per-document breakdown is provided in the Supplementary Table 2. Per-patient expression counts ranged from 201 (P4) to 626 (P3), consistent with genuine differences in clinical complexity and document length across the cohort.

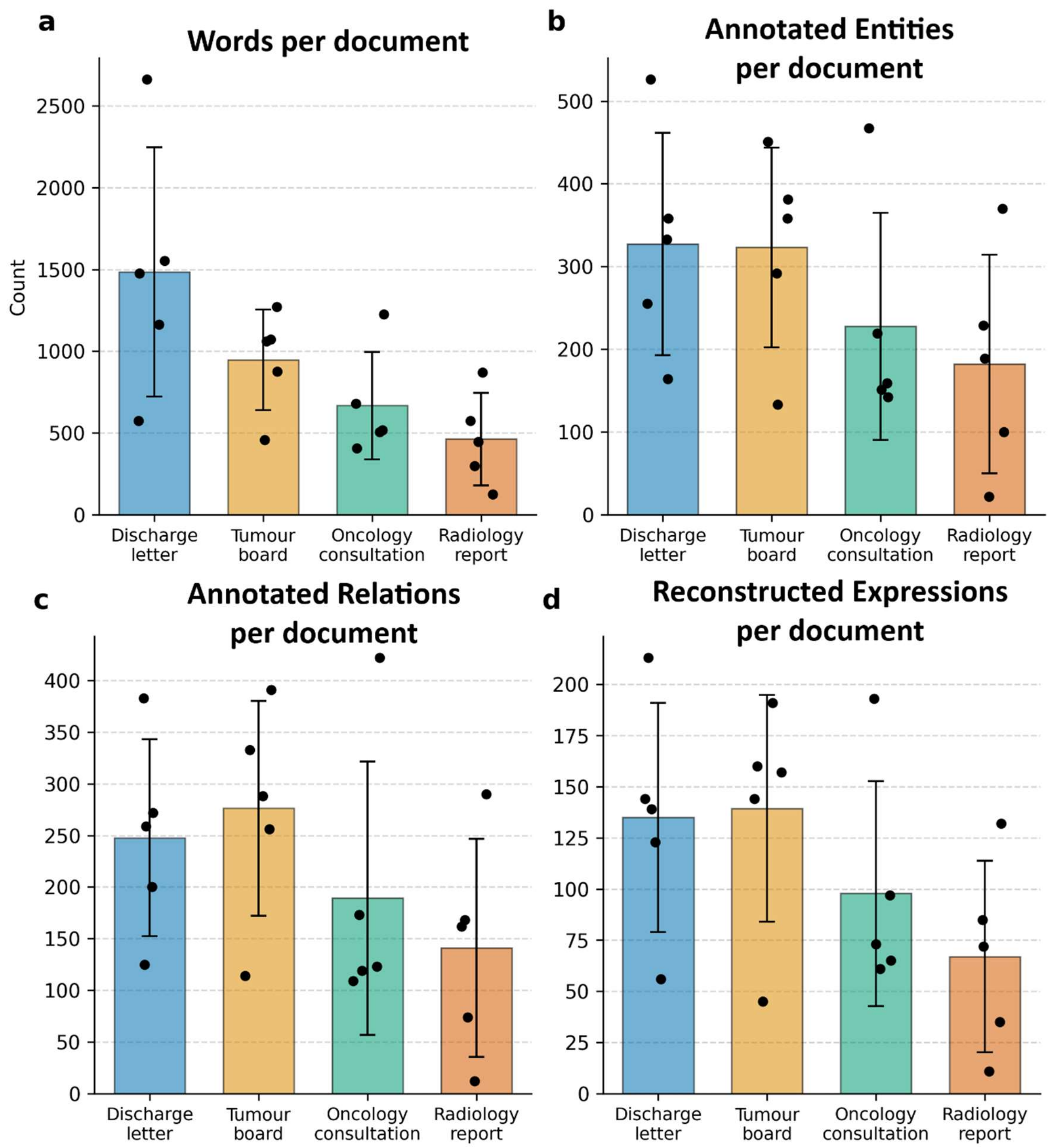

*Figure 7: Mean counts per document (± SD, n=5 patients) for words, annotations, annotated relations, and reconstructed expressions, stratified by document type. Full per-document values are reported in the Supplementary Table 2. Dots: individual documents.*

## Expression density

Across all 20 documents, the pipeline generated a mean of 12.9 (SD 2.9) expressions per 100 words, computed per document. Oncology consultation letters (14.45 per 100 words) showed the highest density, followed closely by tumour board reports (14.21) and radiology reports (13.35). Discharge letters were the consistent exception (9.41), reflecting the greater proportion of narrative prose relative to codeable clinical content in that format.

## Annotation structure

The mean number of annotations per expression was 2.41, indicating that the typical reconstructed expression integrates multiple independently labelled elements rather than reflecting a one-to-one mapping between annotation and output. Annotation spans covered a mean of 1.35 words. The longest single annotation spanned 10 words: *"effet de masse sur les structures de la ligne médiane"* was normalised to the single SNOMED CT concept “299728009 | Midline shift of brain (finding) |”.

## Relation use and class distribution

Annotated Relations, the explicitly labelled directional links between annotation spans that the taxonomy consumes, numbered 3,855 across the corpus, corresponding to a mean of 1.76 annotated relations per reconstructed expression. The remaining 418 annotated relations do not contribute to the represented concepts: about 300 are dates (date, start and end dates), which the pipeline attaches to the reconstructed expressions as temporal metadata rather than as attribute refinements; 91 are grouping relations, which the split mechanism uses to generate one expression per coordinated element; and about 30 are not used by the rule set of the version analysed here (simultaneity links, ordinals and isolated labels; Supplementary Note 2). The distribution across relation classes was uneven: Class III (governing-concept instantiation) accounted for the largest share (34.4%), followed by Class V (narrative connection, 26.3%) and Class I (intrinsic attribute absorption, 22.5%), while Classes IV and II were less frequent (11.5% and 5.3% respectively). Class III’s prominence reflects its role in instantiating the governing clinical concept for most annotated content; the remaining classes describe how that content is refined before this step (Classes I–II) or connected into broader narrative structure afterward (Classes IV–V) (Figure 8).

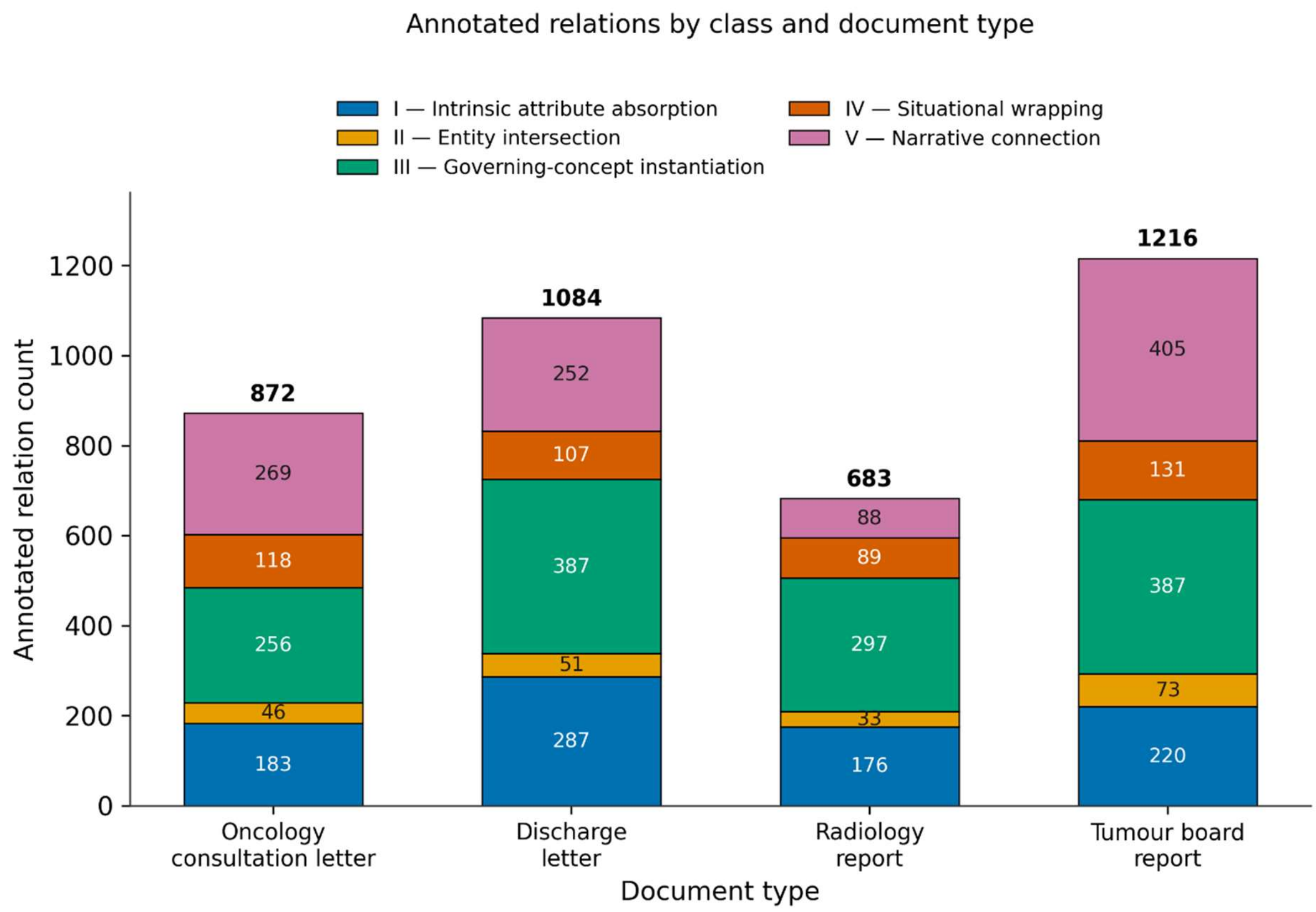

*Figure 8: Distribution of annotated relations by class across the four document types (aggregated across n=5 patients). Class labels follow the taxonomy defined in Table 1.*

## From annotation to post-coordination

The flat relational structure of the source annotation does not map one-to-one onto the post-coordinated output: reconstruction expands and restructures annotated relations into a richer expression graph. Across the corpus, reconstruction produced 7,574 relations in the generated expressions compared to 3,855 at the source level, a mean expansion factor of 1.96.

This expansion was not uniform across relation classes. Class III (governing-concept instantiation) showed the greatest expansion (2.66-fold), reflecting the role of governing-concept instantiation in generating multiple attribute refinements from a single annotated relation: one annotated link between a clinical concept and its modifier unfolds into a set of structured attribute-value pairs in the post-coordinated expression, each constituting a distinct reconstructed relation. Class V (narrative connection), by contrast, showed near one-to-one correspondence (1.06-fold), as these relations connect already-reconstructed expressions to one another rather than contributing internal structure to any single one; their role is coordinative rather than compositional. The per-class expansion factors are shown in Figure 9.

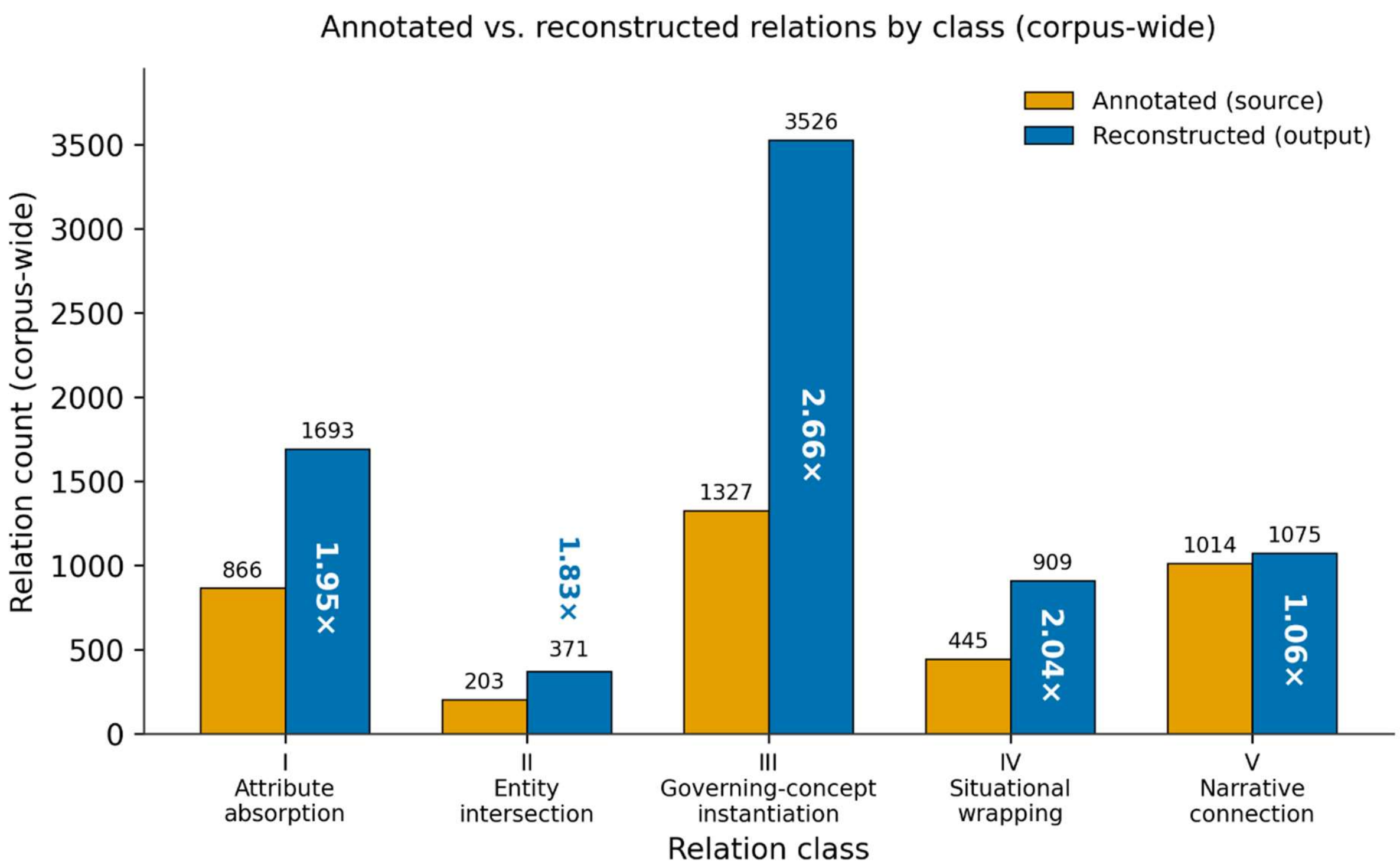

*Figure 9: Annotated relations (source layer) versus reconstructed relations (output layer) by relation class, corpus-wide. The factor shown on each reconstructed bar quantifies the structural expansion introduced by the reconstruction taxonomy for that class.*

## Reconstruction complexity

A typical expression integrates 2.41 source annotations connected by 1.76 annotated relations, which the taxonomy expands into 3.45 reconstructed relations in the final post-coordinated expression, a mean expansion factor of 1.96 (Figure 9). The annotation-to-expression funnel follows a broadly similar shape across document types, though radiology reports show higher counts on both measures (Figure 10).

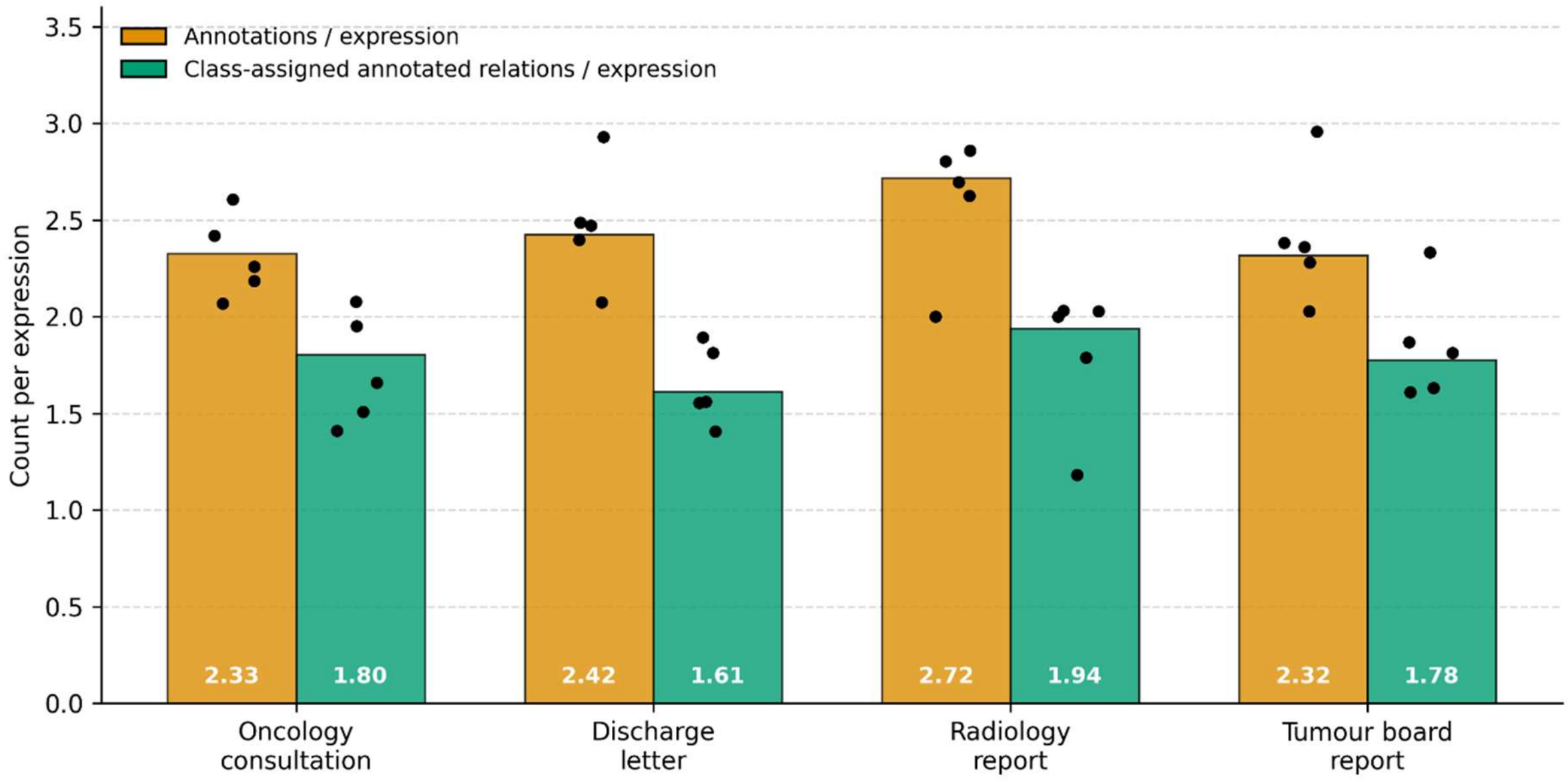

*Figure 10: For each document type, bars show annotations per expression and class-assigned annotated relations per expression, pooled over the five documents of that type; dots show the individual documents.*

## Ontological coverage

Across the full corpus, 1,131 distinct SNOMED CT concepts and 62 distinct NCIt concepts were used. NCIt concepts accounted for 4.0% of all concept assignments, concentrated in clinical areas where SNOMED CT coverage was insufficient: response assessment, molecular and genomic mutations, and the definition of treatment lines and cycles. The relatively low frequency of NCIt assignments reflects the specificity of its role rather than its marginal importance: the content it covers is not otherwise fully representable with SNOMED CT alone, and without it, 210 expressions (9.6% of the corpus) would be either incorrect or unrepresentable.

## Qualitative pipeline verification and error taxonomy

Two expert clinical curators reviewed 153 reconstructed expressions (built from 383 annotated entities and 298 annotated relations) via a purpose-built browser interface linking each element to its source text span, discussing each flagged expression together until its failure cause was agreed upon (rationale for this qualitative review design in Methods, Pipeline verification protocol). Of the 153 expressions, 121 (79.1%) were judged correct and complete (full breakdown in Supplementary Table 3); the 31 expressions flagged by at least one reviewer fell into four recurring failure categories, detailed below.

**Category A: Representational divergence.** The reconstructed expression is semantically valid and clinically defensible but differs from the reviewer's own preferred construction: for example, a different parent relation or normalisation granularity. This does not impair queryability: any expression built from a correctly normalised entity remains reachable by an attribute-based ECL query regardless of which valid compositional choice was made. A system need not resolve every representational ambiguity to produce clinically useful output; enforcing a single canonical construction per type, where required for formal gold-standard evaluation, is a tractable specification change rather than a limitation of the approach.

**Category B: Annotation error.** A number of failures traced to an upstream labelling or normalisation mistake rather than to reconstruction logic: the pipeline correctly executed the annotation as given. Because reconstruction is fully deterministic, every error is traceable to its origin, and in these cases that origin was unambiguously the annotation stage, not the pipeline.

**Category C: Ontological coverage gap.** Few expressions (for example, precise cellular-level pathological observations) had no acceptable supporting concept in either SNOMED CT or NCIt. These do not reflect a limitation of the taxonomy itself, which is ontology-agnostic by design (see Discussion): it operates on any graph-based ontology whose concepts map to entity types and relation classes. Where coverage gaps arise, the appropriate response is targeted ontological extension.

**Category D: Uncovered construction pattern.** A small number of expressions revealed reconstruction paths not yet covered by the current rule set, surfaced by natural-language patterns not previously encountered in the corpus: for example, the absence of a reconstruction path for observable entity orphans, which left certain inter-Assertion relations unpromotable. Each was traceable to a specific missing rule and resolved by extending the rule set; comparable gaps may surface when the pipeline is applied to a new corpus, reflecting the need for iterative rule-set adaptation rather than a defect in the deterministic reconstruction logic. Following correction of the identified rule-set gaps, this category's incidence across the analysed corpus is now zero; residual risk applies only to as-yet-unencountered edge case patterns in future corpora.

**Structural edge cases in ECL representation.** Two constraints inherent to ECL's design surfaced during the reviewing process and mark the outer boundary of what the pipeline can represent, independent of the categories above. First, ECL has no mechanism for placing uncertainty on a relation itself rather than on one of its concepts: when a causal link between an asserted finding and an asserted agent is uncertain (e.g. a reaction *probably due to* a drug), the only available proxy is to contextualise the finding node, which is not semantically equivalent. Second, when the clinically intended focus is embedded within a compound phrase rather than its head (as in *"history of renal tumour, which was excised"*) the pipeline currently assigns the focus of the excision to the assembled antecedent situation rather than the tumour itself, and cannot yet distinguish this from structurally similar but semantically different constructions such as "*biopsy for suspected tumour*". Both edge cases still produce queryable expressions and are documented here as candidates for future refinement.

## ECL queryability

We present illustrative ECL queries here, within Results, to demonstrate the practical query capability whose motivation was introduced above: unified ECL-based cohort discovery across structured and unstructured sources. The post-coordinated expressions generated by the taxonomy are formally representable in Expression Constraint Language, providing a standard semantic specification for retrieval queries against the knowledge graph. In practice, queries are executed via a Cypher[32] translation against the Neo4j[33] graph, with explicit handling of subsumption and IS_A relations. To illustrate the range of clinically meaningful queries this enables, a synthetic patient record from the FRACCO[34] dataset annotated using the same pipeline as the study corpus was used to demonstrate three queries of increasing complexity (Figures 11-13).

**Query 1: Document navigation within a patient record**. A clinician reviewing a complex oncology record wants to surface all documents and passages discussing lung malignancy, without reading the full record. The query filters on finding site and associated morphology (Figure 11):

```
<< 363358000 |Malignant neoplasm of lung (disorder)|  OR
<< 404684003 |Clinical finding (finding)| :
    116676008 |Associated morphology (attribute)| = << 1240414004 |Malignant neoplasm (morphologic abnormality)|
    AND 363698007 |Finding site (attribute)| = << 39607008 |Lung structure (body structure)|
```

*Figure 11: ECL query retrieving lung malignancies, whether they were annotated as a pre-coordinated concepts or post-coordinated expressions*

This query is translated into Cypher to retrieve the corresponding events from the Neo4j graph, from which the originating text spans are recovered for human review. Applied to our synthetic document, it returns 5 expressions, grounded in passages such as the one shown in Figure 14 (panel Q1).

The two-branch structure is needed wherever a primitive pre-coordinated concept may exist for the retrieved content. A phrase normalised directly to such a concept (here 363358000 |Malignant neoplasm of lung|) is stored as a single node, reached only by the first branch; the same content reconstructed as a post-coordinated expression is reached by the second. Ideally, a sufficiently expressive post-coordinated query would subsume the pre-coordinated form via IS_A inference, making one branch sufficient, but given the shortcomings of pre-coordinated expressions documented in Figure 1, covering both explicitly is a best-practice recommendation for broader applicability, rather than a feature exercised within the present dataset: annotation here was performed exclusively via post-coordination, so this dual-branch design does not apply to our own corpus, but including it makes the query strategy directly usable against other datasets or institutions where legacy pre-coordinated concepts may still be present. Queries 2 and 3 use the post-coordinated form alone, since no pre-coordinate exists for their composite content in the present corpus; for other corpora, any query where one plausibly could, should include both branches.

**Query 2: Cohort identification with attribute filtering.** A clinical researcher wants to identify patients in whom metastatic disease was explicitly ruled out by MRI, a cohort that is invisible to standard pre-coordinated coding, which in the best cases captures what was found but not how it was found or what was excluded by a specific investigation. The query combines a Class V inter-event attribute (*finding method*) with a Class IV contextual wrapper (*finding context*) that represents the negation (Figure 12):

```
<< 243796009 |Situation with explicit context (situation)| :
   246090004 |Associated finding (attribute)| = << 404684003 |Clinical finding (finding)| :
       116676008 |Associated morphology (attribute)| = << 14799000 |Neoplasm, metastatic (morphologic abnormality)|
       AND 418775008 |Finding method (attribute)| = << 113091000 |Magnetic resonance imaging (procedure)|
   AND 408729009 |Finding context (attribute)| = << 410516002 |Known absent (qualifier value)|
```

*Figure 12: ECL query retrieving situations where a metastasis was excluded using MRI*

In the graph, finding method is a Class V link connecting two Assertions (the MRI procedure and the metastasis finding) within a Story. Because ECL has no construct for inter-Assertion links, the query expresses it as a refinement on the associated finding, which the Cypher translation resolves by traversing the corresponding Story edge. This query returns an expression grounded in a passage shown in Figure 14 (Q2), where MRI explicitly ruled out metastatic involvement in the pancreas. A retrospective cohort defined this way would otherwise require exhaustive manual review of free-text radiology and MDT reports; the reconstruction taxonomy enables queries that reduce that to a targeted review of a computable candidate set.

**Query 3: Cross-ontology retrieval.** A researcher wants to retrieve all expressions in which a disease response assessment was recorded, a concept not adequately represented in SNOMED CT and therefore encoded using NCIt terminology in our dual-ontology design. Following the external ontology query representation proposed by Gaudet-Blavignac et al.[9], the query becomes (Figure 13):

```
<< 404684003 |Clinical finding (finding)| AND << ~NCIt~Disease Response (Code - C50995)~
OR
<< 404684003 |Clinical finding (finding)| :
    263502005 |Clinical course (attribute)| = << ~NCIt~Disease Response (Code - C50995)~
```

*Figure 13: ECL query retrieving expressions of disease progression (corresponding to the NCIt "Disease Response" family) whether as a clinical course of an existing disorder or as a standalone finding, as it is sometimes expressed in oncology reports.*

The two-branch structure here reflects the annotation design rather than the pre/post-coordination duality: a disease response concept may appear either as a clinical course attribute of an associated finding, or as a standalone finding in its own right when no parent event is present in the text. Both forms are queried to ensure complete retrieval.

This query returns 7 expressions from the synthetic record, spanning two expression subtypes: direct disease response events and clinical course attributes of associated findings. Retrieved passages include references to disease progression at specific anatomical sites as well as findings such as a standalone disease progression, as shown in Figure 14 (Q3). The full Cypher translations in the context of our experimental setup for these queries are provided on the repository (see Code availability).

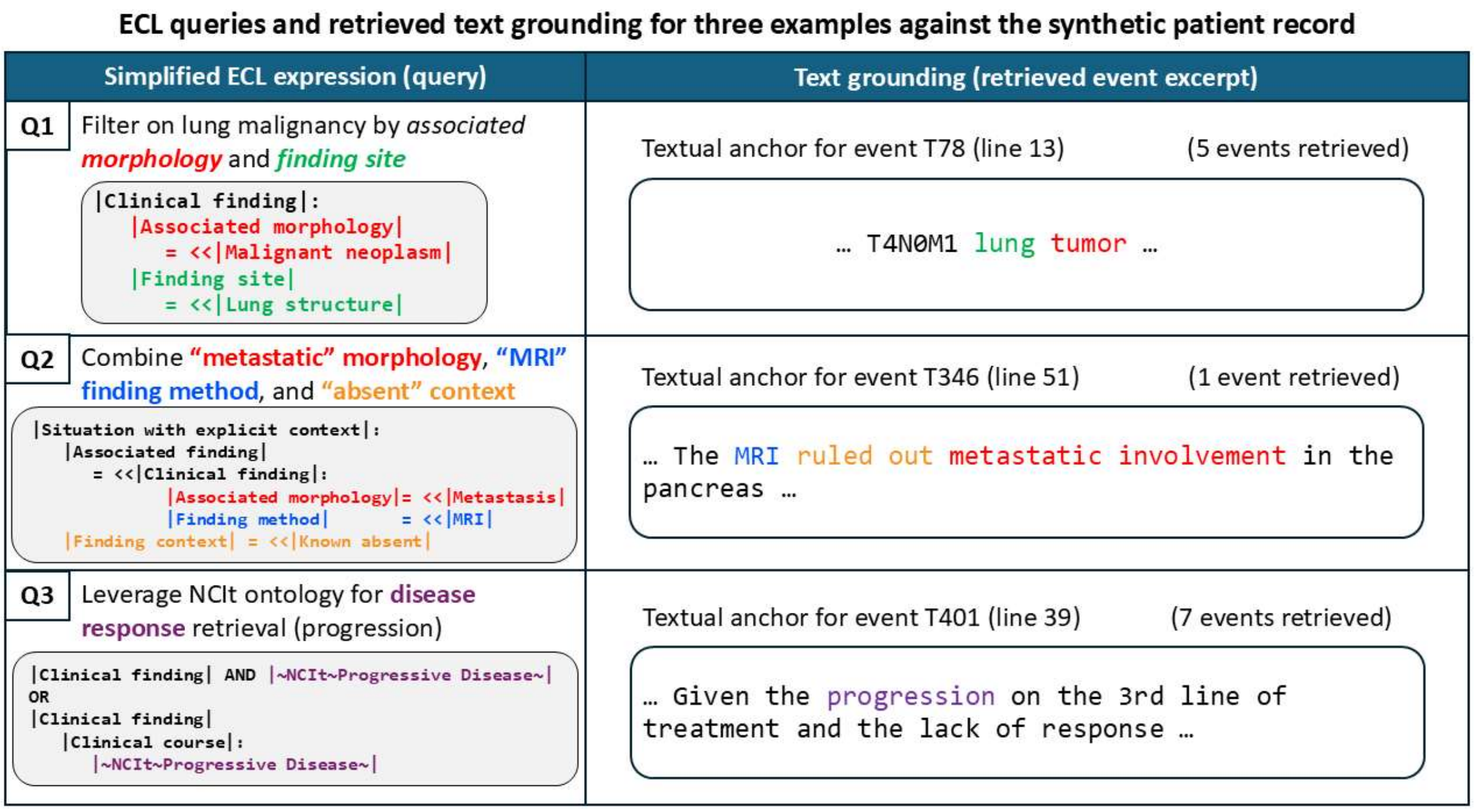

*Figure 14: Retrieved events are grounded to annotated spans; surrounding tokens are shown for readability. Highlighted spans indicate the annotated entity corresponding to the focus concept or attribute of the retrieved expression. Panels Q1, Q2 and Q3 correspond to the queries in Figures 11, 12 and 13, respectively.*

# Discussion

The central finding of this study is that, although prior work has shown entity extraction and normalisation can be performed with reasonable accuracy[27,28], they capture only which concepts a document mentions, not how those concepts combine into a single structured expression. A flat list of mentioned concepts can still be queried for presence or absence; what it cannot do is represent a composed clinical assertion: relation structure is the missing, binding constraint. The relation taxonomy — a classification system defining how annotated entities may be combined, distinct from the annotation process that supplies its raw material — provides the construction grammar that converts that material into formally queryable knowledge. The three structural failures of pre-coordinated concept lookup identified in the introduction are precisely what the taxonomy circumvents. When no pre-coordinated concept exists for a clinical combination, post-coordination assembles the expression from atomic components that do, making the possible absence of a compound pre-coordinated SNOMED concept irrelevant as queries traverse the assembled semantic structure[26]. When a target concept is primitive, the assembled expression bypasses it, grounding queryability in explicitly asserted attribute edges rather than inherited ontological definitions[23]. When a fully defined concept exists but its attribute values are misaligned with expected query paths, atomic assembly selects values for their position in the query-relevant hierarchy rather than inheriting them from a curated compound concept[24]. In each case, the output is queryable by construction, through the same ECL mechanisms that operate on existing structured data[25].

The five construction Rules derive from the logical structure of SNOMED CT post-coordination[23], and from the compositional grammar of clinical language more broadly[9]. These Rules are not specific to oncology: the reconstruction logic operates on relation class assignments and entity mappings provided at annotation time and extending it to a new clinical domain requires only that the entity type vocabulary and the Rule III attribute instances be configured appropriately for that domain. The integration of NCIt as a supplementary normalisation target illustrates the same principle of configurability: NCIt concepts participate in expression assembly identically to SNOMED CT concepts. The principal operational constraint encountered in this study, namely that ECL hierarchy traversal operates over the SNOMED CT graph alone, is a property of the current experimental setup rather than of the approach.

The more consequential implication of the Relation Classes concerns the automation of clinical NLP. The path toward computationally accessible clinical records has been framed, for the better part of two decades, as a problem of entity extraction: identify the relevant spans, normalise them to terminology codes, and the knowledge representation follows. What our graph reconstruction framework makes explicit is that this framing is incomplete. Entity extraction and terminology normalisation are necessary preconditions, but the step that allows to convert a set of normalised entities into a formally queryable clinical fact is drawing and labelling the relations between them, each relation type already bound to a specific and non-substitutable construction rule. Reframing clinical NLP as a relation-first problem changes what evaluation targets and training objectives are appropriate. A system that correctly identifies entities and normalises them to SNOMED CT concepts but produces untyped or arbitrarily typed relations between them cannot be used as input to a post-coordination pipeline. A system designed to identify entities correctly and label the relations between them, using the fixed relation vocabulary defined here, produces output that is directly consumable by the reconstruction algorithm. The annotated corpus described in this study constitutes an initial resource for that training and evaluation task. Large language models, which have demonstrated strong zero-shot and fine-tuned performance on structured clinical information extraction, are a natural candidate for the extraction layer, given their capacity to capture the contextual and syntactic patterns that determine relation class membership[4,11,35].

Several limitations warrant consideration. The primary constraint for actual deployment is the manual annotation described here, which does not scale to large corpora without substantial resource investment[36,37]. This reflects the current development stage of the methodology rather than an intrinsic constraint of the approach: because extraction logic and reconstruction logic are explicitly decoupled, automating the extraction layer requires no modification to the downstream assembly pipeline. The class-bound relation vocabulary provides the formal evaluation target that such automation requires. A further limitation concerns expressions for which no explicit focus concept is available in the source text; these are assigned a high-level SNOMED root, preserving attribute-value content at the cost of hierarchical specificity. For attribute-based cohort discovery[30,31] this trade-off is acceptable, but resolving it fully would require a contextual inference step beyond the scope of the current pipeline. The corpus comprised five patients from a single melanoma cohort, purposefully selected to exercise the full range of relation classes and construction patterns in the taxonomy. Because the construction Rules derive from the formal structure of SNOMED CT post-coordination rather than from statistical patterns in the corpus, its validity as a construction system is not contingent on corpus size in the way an empirical model's would be. What application across other clinical domains, institutions, and documentation styles will determine is not whether the Rules generalise, but what entity type extensions and attribute configurations are required to cover the construction patterns specific to those contexts. The same distinction applies to language: the corpus analysed here is in French, but because the Rules operate over normalised SNOMED CT and NCIt concepts rather than surface text, the resulting graph representation is itself language-agnostic; the five relation Classes are expected to generalise to languages with comparable clinical narrative conventions, though the specific relation type and vocabulary bindings instantiating each Class were developed in consultation with this institution's practice and may require local adaptation elsewhere. A related boundary concerns narrative scope: Stories (Rule V) are limited to the document in which they are constructed. Connecting Stories across a patient's multiple documents is a natural next step, but one that moves from representation (structuring what a document explicitly states) toward interpretation (inferring relationships the source text does not itself assert) and is treated as out of scope for the present deterministic pipeline.

A decade of clinical NLP research has established entity extraction and terminology normalisation as necessary components of clinical text formalisation. The present work shows that they are not sufficient: a typical post-coordinated expression required an average of 3.45 reconstructed relations from just 1.76 annotated relations (see Reconstruction complexity). The missing component is a principled class-bound relation vocabulary and associated construction rules that operationalise the Extended Grammar directly from free text. The five Rules presented here provide that construction grammar: a domain-invariant specification that converts extracted entities into attribute-rich, ECL-queryable expressions and establishes a formal evaluation target for the next generation of automated clinical relation extraction systems. By grounding free-text-derived knowledge in the same ontological structures as existing structured clinical data, the methodology makes unified cohort discovery across structured and unstructured sources operationally tractable.

The broader significance of this work for digital medicine lies not in any single query it enables, but in what its output is: a standing, reusable representation rather than a one-off answer. Cohort identification for clinical research currently depends on either incomplete structured codes or exhaustive manual review, and both must be repeated whenever the research question changes. A knowledge graph assembled by this method is instead persistent: once built, it supports arbitrary future ECL queries over the same patient population without further manual review, on equal footing with structured EHR data. Because the relation Classes and the associated construction Rules are ontology-agnostic and configured, rather than retrained, for a new domain, this persistence is not bound to any single clinical specialty: the same construction grammar renders free-text documentation queryable wherever formal attribute-based retrieval is needed, whether the question concerns which patients had a complication excluded by a specific investigation, which experienced a defined adverse event on a given medication, or which meet a compound eligibility criterion for a trial. What this makes newly possible, at scale, is treating free-text clinical narrative itself as an interactable resource: not a document to be re-read each time a new question is asked, but a source that can be queried directly, repeatedly, and for questions not yet anticipated at the time it was written.

# Methods

## Study context and corpus

As introduced above, this work aims to extend an institutional knowledge-graph architecture in which structured EHR fields are represented and manually normalised to SNOMED CT concepts and queried using Expression Constraint Language (ECL).

This study was approved by the Geneva Cantonal Commission for Research Ethics (Commission Cantonale d'Éthique de la Recherche, CCER; BASEC ID 2025-00719), which granted an exception to the requirement for documented general consent given the minimal risk to patients beyond data confidentiality; patients with a documented refusal of general consent were excluded from the cohort.

Patient selection proceeded in three stages. Eligibility was first restricted to patients covered by the aforementioned ethics protocol. From the approximately 100 eligible patients, the study's data manager selected five with sufficiently complete documentation and extraction variables to support the study, independent of narrative content. For each selected patient, the most recent occurrence of the four document types routinely used for institutional clinical curation were extracted: discharge letter, oncology radiology report, tumour board report, and oncology consultation report. This sampling strategy was designed to ensure that all major relation classes and construction patterns in the taxonomy were exercised within the corpus. Documents were processed through an in-house PDF extraction interface to isolate clinically relevant narrative text, followed by automated de-identification. All 20 documents were manually reviewed and, where necessary, corrected to ensure complete removal of identifying information prior to annotation. Embedded tables, generated from structured EHR fields already captured in the existing graph, were excluded. A synthetic example of the documentation is provided in Supplementary Note 1 and on the repository (see Code availability) to ensure reproducibility.

## Annotation framework

Cleaned documents were imported into BRAT for manual annotation following a defined schema (Supplementary Note 2; schema files on the repository). Annotation in BRAT captures both entity spans with an assigned normalisation code, and the relations between them: relation type and directionality are recorded by the annotator at annotation time. Concepts were normalised to SNOMED CT International Edition (release 20250501) and NCIt (version 26.05d). Entity types align with SNOMED CT; NCIt extends coverage to oncology-specific concepts not well represented in SNOMED CT, following the integration rules of Gaudet-Blavignac et al.[9] The annotation strategy is *atomic*: complex noun phrases are decomposed into the smallest spans independently mappable to an ontology concept. For example, "melanoma of the left leg" yields three separate entities ("melanoma," "left," and "leg") whose relations are then captured explicitly, enabling assembly of a fully defined, attribute-rich post-coordinated expression rather than a lookup against a potentially primitive compound concept. Full entity type definitions and relation domain/range constraints enforced within BRAT are provided in Supplementary Note 2 and on the repository. Annotation was performed by a single expert annotator. Span, normalisation, and relation errors introduced at this stage were expected to surface during expert validation, and were indeed identified and categorised as a distinct failure class (Category B) in the validation results.

The five relation classes and their constituent relation types were developed iteratively over the 20-document corpus: an initial pass in which taxonomy development and annotation proceeded jointly, document by document, followed by a second pass reconciling earlier documents to the final taxonomy. Development was led in consultation with the institution's post-coordination experts to keep relation class definitions aligned with the institution's established practice. The running example sentence (Figures 2 and 3), which deploys all five classes at once, illustrates the most complex construction supported by the compositional grammar formalised by Gaudet-Blavignac et al.[9]

## Pipeline verification protocol

Verification was performed at the level of the reconstructed expression and covered every expression of two complete documents, an oncology consultation letter (97 expressions) and a discharge letter (56 expressions): 153 expressions built from 383 annotated entities and 298 annotated relations. Because each expression is assembled deterministically from its annotations, verifying an expression also verifies the entities, normalisations and relations it was built from. Review proceeded in two passes. In a first pass, two expert clinical curators examined each reconstructed SNOMED Compositional Grammar expression alongside its originating source text, using a purpose-built browser interface with hover-linked cross-referencing between expression elements and annotated text spans. For each expression, they assessed whether the generated output correctly and completely represented the clinical content of the source text, and recorded a written justification for any expression judged incorrect or incomplete; both curators reviewed the consultation letter, and one reviewed the discharge letter. In a second pass, the two curators and the annotator reviewed every flagged expression together until its cause was identified and agreed upon by consensus. An expression was counted as initially correct and complete when no reviewer flagged it. Both curators have applied manual post-coordination to this institution's structured EHR fields for approximately five years; assessment drew on this standing clinical expertise, including the recognition that a given clinical meaning can admit more than one valid post-coordinated formulation, which makes raw inter-annotator agreement a poor measure of correctness here.

Because the reconstruction pipeline is fully deterministic and rule-based, review proceeded as joint expert verification rather than independent-rater annotation: the consensus pass worked in the manner of a code review, tracing each flagged expression back to its cause. This follows from what is being validated: where a system's behaviour is probabilistic, aggregate accuracy quantifies a genuine uncertainty in its output; where behaviour is deterministic, an error has a single identifiable cause in the pipeline logic or its input, and the relevant question is what shape that error takes rather than how often it occurs. The goal of the review was accordingly to establish whether the pipeline behaves as intended across the range of clinical constructions present in the corpus, and to characterise any systematic failure modes prior to broader application. This scope was sufficient for this purpose: reviewers observed convergence in failure modes across the flagged expressions, consistent across the four categories reported in Results (Qualitative pipeline verification and error taxonomy).

## Pipeline implementation

The reconstruction pipeline, UNIGRAPHE, is implemented in Python (3.10) and operates against a Neo4j graph database (Community Edition, 2026.06.0), deployed via Docker Compose alongside a Python backend service. Annotation was performed in BRAT (v1.3)[38]; BRAT annotation files (.ann format) are consumed directly as pipeline input. For each document, Annotated Entity spans and Relations are parsed from the .ann file and passed to the reconstruction algorithm, which executes the five Rules sequentially, generating nodes and relationships to Neo4j via the Bolt protocol using the official Python driver. Node types (Annotated Entity, Qualified Entity, Composite Entity, Governing Concept, Situation, Story) are created and linked progressively across pipeline steps, in accordance with the eligibility and Assertion-role rules defined in Table 1; each node retains an explicit provenance edge to its originating text span. All source files, Cypher query templates, and configuration files are available on the dedicated repository (see Code availability).

An Annotated Relation is fixed at annotation time and always connects two Annotated Entities, and is assigned to a Relation class by its label; this classification never changes over the course of reconstruction. When a Class's corresponding Rule processes an Annotated Relation, each of its endpoints is resolved to whatever node currently represents it in the graph: that node may be the original Annotated Entity itself, or a Qualified or Composite Entity already built from it. Multiple Rule applications can contribute to this resolution, since Rule I and Rule II may apply repeatedly to the same Annotated Entity . The step's actual input is the resolved node, not the Annotated relation's literal endpoint. Applying a rule also creates a Reconstructed Relation: a new relation, typically sharing the Annotated Relation's label, attached to the resulting node.

## Statistics and reproducibility

All corpus statistics are descriptive. Counts are reported per document (n = 20 documents from n = 5 patients, one document of each of the four types per patient) and summarised as means with standard deviations (SD) across documents or patients, as indicated; ± denotes SD throughout. Expression density was computed per document and then averaged. No inferential statistical tests were performed. Qualitative verification covered all 153 expressions of two complete documents. Because the reconstruction pipeline is deterministic, re-running it on the same annotation files reproduces the reported graphs exactly.

## Manuscript preparation

Sections of this manuscript were revised with the assistance of Claude Sonnet 4.6 (Anthropic), a large language model. All content was reviewed, verified, and approved by the authors, who take full responsibility for the accuracy and integrity of the work as presented.

## Data Availability

The synthetic patient document and associated annotations used to demonstrate ECL queryability are available on the dedicated repository https://github.com/SimedDataTeam/unigraphe. The real-world oncology corpus was derived from patient records held under BASEC ID 2025-00719 and cannot be made publicly available due to patient confidentiality constraints, particularly the high reidentification risk associated with free text; for the same reason, it cannot be shared on request. Aggregate per-document statistics are provided in Supplementary Table 1.

## Code Availability

The knowledge graph reconstruction algorithm, including Cypher query templates and JSON configuration files as well as the BRAT annotation schema files and the expression viewer and corresponding synthetic file are also available on the dedicated repository (https://github.com/SimedDataTeam/unigraphe).

## Acknowledgments

This work was funded by GTOP (*Geneva Translational Oncology Programme*) and the NOVA foundation (*Fondation pour l'Innovation sur le Cancer et la Biologie*, https://www.fondation-cancer.ch/), with help from the Leenards Foundation (https://leenaards.ch/) (J.-P.G), with a grant from the Private Foundation of the Geneva University Hospitals (https://www.fondationhug.org/). This project benefited from infrastructure provided by the HUG. Open access funding provided by University of Geneva. The funders played no role in study design, data collection, analysis and interpretation of data, or the writing of this manuscript.

## Author Contributions

J.P. developed the taxonomy and algorithmic resolution of the presented methodology, based on C.L., C.G-B and J.E. previous work on grammatical rules for post-coordination[9]. Manual annotation and subsequent application of the pipeline was conducted by J.P. C.G-B and J.E. reviewed the verification dataset and discussed all cases with J.P to reach an accord. J-P.G. and P.C participated on the extraction and curation work, while C.D. provided technical support throughout the project. O.M., C.L., M.C., and M.B. supervised the entire process and gave the final approval. Writing of the article was conducted by J.P. with contributions from C.G-B., J.E., and M.C. All authors reviewed and approved the final manuscript before submission.

## Competing Interests

The authors declare no competing interests.